# COMPLEX SPECTROGRAM ENHANCEMENT BY CONVOLUTIONAL NEURAL NETWORK WITH MULTI-METRICS LEARNING

*Szu-Wei Fu [12], Ting-yao Hu[3], Yu Tsao[1], Xugang Lu[4]*

[1] Research Center for Information Technology Innovation, Academia Sinica, Taipei, Taiwan
[2] Department of Computer Science and Information Engineering, National Taiwan University, Taipei, Taiwan
[3] Department of Computer Science, Carnegie Mellon University, Pittsburg, PA, USA.
[4] National Institute of Information and Communications Technology, Kyoto, Japan

## ABSTRACT

This paper aims to address two issues existing in the current speech enhancement methods: 1) the difficulty of phase estimations; 2) a single objective function cannot consider multiple metrics simultaneously. To solve the first problem, we propose a novel convolutional neural network (CNN) model for complex spectrogram enhancement, namely estimating clean real and imaginary (RI) spectrograms from noisy ones. The reconstructed RI spectrograms are directly used to synthesize enhanced speech waveforms. In addition, since log-power spectrogram (LPS) can be represented as a function of RI spectrograms, its reconstruction is also considered as another target. Thus a unified objective function, which combines these two targets (reconstruction of RI spectrograms and LPS), is equivalent to simultaneously optimizing two commonly used objective metrics: segmental signal-to-noise ratio (SSNR) and log-spectral distortion (LSD). Therefore, the learning process is called multi-metrics learning (MML). Experimental results confirm the effectiveness of the proposed CNN with RI spectrograms and MML in terms of improved standardized evaluation metrics on a speech enhancement task.

*Index Terms*—Convolutional neural network, complex spectrogram, speech enhancement, phase processing, multi-objective learning

## 1. INTRODUCTION

Recently, various types of deep-learning-based denoising models have been proposed and extensively investigated [1-12]. They have demonstrated superior ability to model the non-linear relationship between noisy and clean speech compared to traditional speech enhancement models. However, most existing denoising models focus only on processing the magnitude spectrogram (e.g., log-power spectrogram, LPS) leaving phase in its original noisy condition. This may be because there is no clear structure in the phase spectrogram, which makes estimating clean phase from noisy phase extremely difficult [13]. On the other hand, some researches have shown the importance of phase when spectrograms are resynthesized back into time-domain waveforms. Roux [14] demonstrated that when the inconsistency between magnitude and phase spectrograms is maximized, the same magnitude spectrogram can lead to extremely diverse resynthesized sounds, depending on the phase with which it is combined. Paliwal *et al.* [15] confirmed the importance of phase for perceptual quality in speech enhancement, especially when window overlap and length of the Fourier transform are increased.

To further improve the performance of speech enhancement, phase information is considered in some up-to-date research [13, 16-19]. For time-domain signal reconstruction, Wang *et al.* [18] proposed a deep neural networks (DNN) model which tries to learn an optimal masking function given the noisy phase. Williamson *et al.* [13, 19] found that the structures in real and imaginary (RI) spectrograms are similar to that of magnitude spectrograms. Therefore, they employed a DNN for estimating the complex ratio mask (cRM) from a set of complementary features, and thus magnitude and phase can be jointly enhanced. The quality of the cRM enhanced speech is improved compared to the ideal ratio mask (IRM) based model.

In this paper, we estimate clean RI spectrograms directly from noisy ones instead of complementary features (e.g., amplitude modulation spectrogram, relative spectral transform and perceptual linear prediction, etc.) used in [13]. To efficiently exploit the relation between RI spectrograms, they are treated as different input channels in the proposed convolutional neural network (CNN) model.

Since the goal of speech enhancement is to improve the intelligibility and quality of a noisy speech [20], several objective metrics have to be applied to evaluate the performance in different aspects. For example, segmental signal-to-noise ratio (SSNR in dB) measure the signal difference in time domain, and log-spectral distortion (LSD in dB) [21] measure the spectrogram difference. Because the outputs of the proposed CNN are RI spectrograms, which do not loss any information from raw waveform, other signal representation forms (e.g., waveform, log power spectrum) can be derived from them. Using this characteristic, several metrics can also be optimized simultaneously by including them into the objective function of our CNN. Each target corresponds to a metric; hence, the learning process is referred to as multi-metrics learning (MML) in this paper. Unlike a usual multi-objective optimization problem [22], the targets in MML do not conflict with each other, which implies that there are no serious trade-offs between different metrics.

## 2. NOISY PHASE

For DNN-based speech enhancement, the noisy and clean speech signals are usually first converted into the frequency domain to extract their LPS as input features and output targets, respectively [1]. The enhanced signal in the time domain can be synthesized from the combination of its enhanced LPS and phase information, which is borrowed from the original noisy speech. Figure 1 presents an example of clean magnitude and phase spectrograms (top) and thresholded phase difference between clean and noisy speech under high and low SNR conditions (bottom). From Fig. 1, we can note that using the noisy phase information may not cause serious problems in high SNR conditions since the noisy phase is similar to the clean phase, even in high-energy regions (bottom-left of Fig. 1). To briefly explain the reason, the noisy phase in a time-frequency (T-F) unit is defined as $arctan2(N_i, N_r)$, where $N_i$ and $N_r$ are the imaginary and real parts of noisy complex spectrogram, respectively, and $arctan2$ is similar to the arc tangent of $N_i/N_r$, except that the signs of both arguments are considered to determine the appropriate quadrant [23]. Here, the expression of phase is simplified as follows:

$$arctan\frac{N_i}{N_r} = arctan\frac{S_i + n_i}{S_r + n_r} \quad (1)$$

where $S_i$ and $S_r$, $n_i$ and $n_r$ are imaginary and real parts of speech and noise, respectively. When the SNR of noisy speech and the speech energy in the T-F unit is high enough, that is $|S_i| \gg |n_i|, |S_r| \gg |n_r|$, then the noisy phase in (1) is similar to the clean phase as (2):

$$arctan\frac{N_i}{N_r} \approx arctan\frac{S_i}{S_r} \quad (2)$$

This well explains why the structures in top-left and bottom-left figure of Fig. 1 are similar to each other. However, this is not the case in low SNR conditions where the quality of the synthesized signal with enhanced phase may be considerably improved [13].

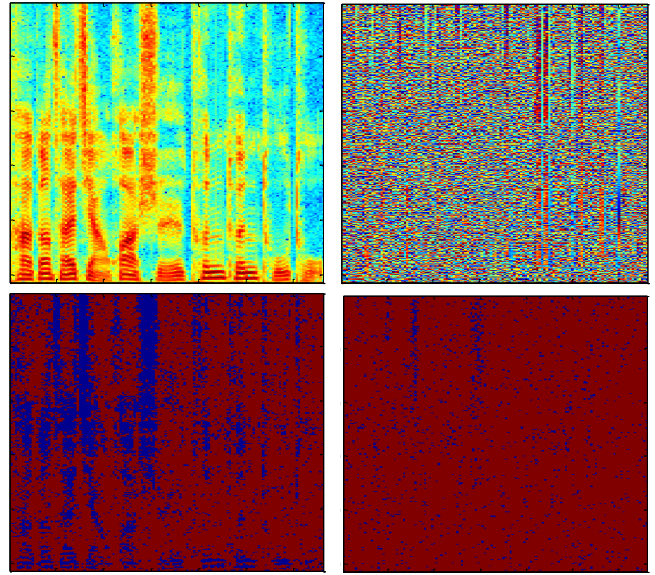

**Fig. 1.** Example of clean magnitude (top-left) and phase (top-right) spectrograms. Phase difference between clean and noisy speech (engine noise) under 12 dB (bottom-left) and -12 dB (bottom-right). Here the regions in blue represent the absolute phase difference is smaller than 0.1.

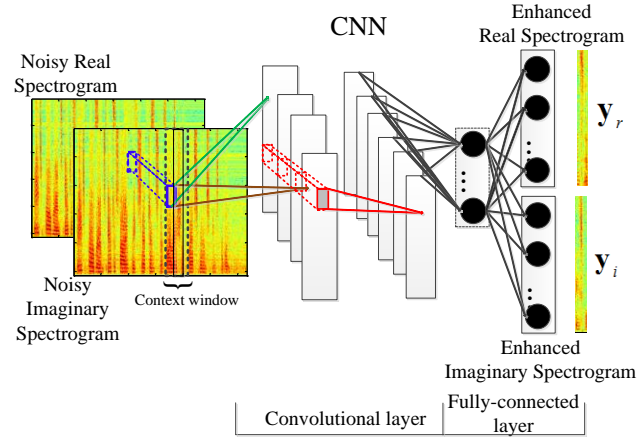

**Fig. 2.** RI spectrograms enhanced by CNN. Real and imaginary spectrograms are treated as different input channels.

## 3. ENHANCEMENT OF RI SPECTROGRAMS BY CNN

One possible way to enhance the phase is to employ a conventional DNN model to estimate clean phase from noisy phase. Due to the lack of structure (as shown in top-right of Fig.1), however, it is difficult for a machine learning model (even for deep learning) to learn the relationship between clean and noisy phase [13]. On the other hand, Williamson *et al.*[19] found that the structures in RI spectrograms are similar to that of magnitude spectrograms.

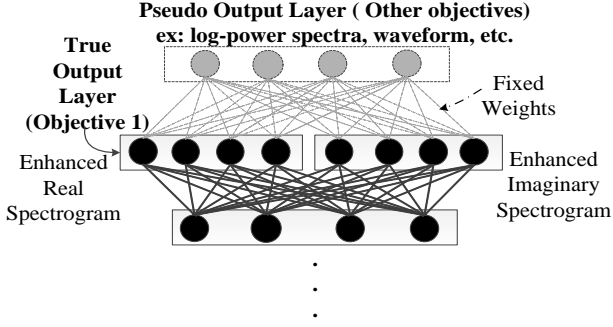

**Fig. 3.** Proposed pseudo network with pseudo hidden and output layer(s).

Based on this observation, Fig. 2 presents the proposed CNN structure for speech enhancement using RI spectrograms. Rather than processing the phase directly, the network aims to estimate clean RI spectrograms from noisy ones. By the definition of phase: $y_p = \text{atan2}(y_i, y_r)$, where $y_p, y_i, y_r$ are enhanced phase, imaginary part, and real part, in a T-F unit, respectively. If the enhanced real and imaginary parts are appropriately processed, the phase part may thereby be enhanced as well. Note that in the proposed CNN structure, real and imaginary spectrograms are treated as different input channels. This is a similar idea for processing RGB channels of a color image in the field of computer vision. Comparing to DNN, which fully connect all inputs in the RI spectrograms, the proposed CNN can concentrate on local pattern, and hence efficiently extract useful features.

The objective function used for the clean RI spectrogram reconstruction can be expressed as follows:

$$O = \sum (\| \hat{\mathbf{y}}_i - \mathbf{y}_i \|_2^2 + \| \hat{\mathbf{y}}_r - \mathbf{y}_r \|_2^2) \\ = \sum \| \hat{\mathbf{y}} - \mathbf{y} \|_2^2 \quad (3)$$

where $\hat{\mathbf{y}}_i, \hat{\mathbf{y}}_r \in R^L$ and $\mathbf{y}_i, \mathbf{y}_r \in R^L$ are clean and enhanced imaginary and real spectrograms, respectively, and $L$ is the dimension of the spectrum. $\hat{\mathbf{y}} = [\hat{\mathbf{y}}_r \ \hat{\mathbf{y}}_i]^T$ and $\mathbf{y} = [\mathbf{y}_r \ \mathbf{y}_i]^T \in R^{2L}$ are the vertically cascaded vectors of the clean and enhanced RI spectrograms, respectively.

## 4. MULTI-METRICS LEARNING

Since the outputs of the proposed network are RI spectrograms, which have the same information amount as raw waveform, other signal representation forms (e.g., waveform, LPS) can be depicted as functions of them.

We will first show that enhancing the RI spectrograms has similar effect as de-noising the waveform directly.

### 4.1. Relation between RI spectrogram and waveform

To directly de-noise a noisy waveform, one possible solution is to apply an objective function to minimize the distance between clean and enhanced waveforms as follow:

$$\| \hat{\mathbf{w}}_y - \mathbf{w}_y \|_2^2 \quad (4)$$

where $\hat{\mathbf{w}}_y, \mathbf{w}_y \in R^{2L-2}$ are the corresponding clean and enhanced waveforms, respectively. This term can also be expressed as a function of $\mathbf{y}$ and $\hat{\mathbf{y}}$ through the inverse discrete Fourier transform (IDFT):

$$\| \hat{\mathbf{w}}_y - \mathbf{w}_y \|_2^2 \\ = \| (\mathbf{CU}_1 \hat{\mathbf{y}}_r - \mathbf{SU}_2 \hat{\mathbf{y}}_i) - (\mathbf{CU}_1 \mathbf{y}_r - \mathbf{SU}_2 \mathbf{y}_i) \|_2^2 \quad (5) \\ = \| \mathbf{F}\hat{\mathbf{y}} - \mathbf{F}\mathbf{y} \|_2^2$$

where $\mathbf{U}_1, \mathbf{U}_2 \in R^{(2L-2) \times L}$ are the matrices used for recovery of the even symmetry of the real part, and the odd symmetry of the imaginary part, respectively. $\mathbf{C}, \mathbf{S} \in R^{(2L-2) \times (2L-2)}$ are the cosine and sine matrices in the IDFT, respectively. $\mathbf{F} \in R^{(2L-2) \times (2L)}$ is defined as:

$$\mathbf{F} = [\ \mathbf{CU}_1 \ -\mathbf{SU}_2\ ] \quad (6)$$

Comparing (3) and (5), it can be observed that the only difference between enhancing RI spectrograms and waveform is the matrix multiplication, $\mathbf{F}$. Since it does not bring any non-linear effects in the back-propagation process, their enhancement results have similar trend. Therefore, optimizing RI spectrograms is related to maximizing SSNR.

### 4.2. Incorporating LPS reconstruction term into the objective function

In this section, we investigate to minimize LSD of enhanced speech by incorporating LPS reconstruction term into the objective function. It can also be expressed as function of $\mathbf{y}$ and $\hat{\mathbf{y}}$ in matrix form as follows:

$$\| \log(\hat{\mathbf{y}}_i^2 + \hat{\mathbf{y}}_r^2) - \log(\mathbf{y}_i^2 + \mathbf{y}_r^2) \|_2^2 \\ = \| \log(\mathbf{P} \times \text{sqr}(\mathbf{I}\hat{\mathbf{y}})) - \log(\mathbf{P} \times \text{sqr}(\mathbf{I}\mathbf{y})) \|_2^2 \quad (7)$$

where $\mathbf{I} \in R^{2L \times 2L}$ is the identity matrix, sqr(.) is the square function, $\mathbf{P} \in R^{L \times 2L}$ is the permutation matrix defined as:

$$\mathbf{P} = [\mathbf{I}_{(L) \times (L)} \ \mathbf{I}_{(L) \times (L)}] \quad (8)$$

From (7), it can be noted that the relation between LPS and RI spectrograms is non-linear. Therefore, this transformation does produce some effects on the enhancement results. Thus, we formulate a unified objective function by combining (3) and (7) as follows:

$$O = \sum \alpha \| \hat{\mathbf{y}} - \mathbf{y} \|_2^2 \\ + \beta \| \log(\mathbf{P} \times \text{sqr}(\mathbf{I}\hat{\mathbf{y}})) - \log(\mathbf{P} \times \text{sqr}(\mathbf{I}\mathbf{y})) \|_2^2 \quad (9)$$

where $\alpha$ and $\beta$ are weighting factors for different target objective functions. Please note that the first term is the original objective function for the RI spectrum used for maximize SSNR. The second term is about log power spectrogram which tries to minimize LSD. Hence, the learning process is called multi-metrics learning in this paper. Although the last term may seem redundant, it actually affects how the enhanced speech approaches the clean speech, which will be discussed later in the experiments. It is not difficult to find that all the terms in (9) are directly related to the output vector **y** and can be expressed as a combination of matrix multiplication and a non-linear function as in a typical neural network. Therefore, the proposed network can be equivalently represented as additional pseudo hidden and output layer(s) with fixed weights, augmenting the true output layer, as shown in Fig. 3. In this paper, we refer this augmentation as the pseudo network for its characteristic and structure. During training, the gradient will pass through the pseudo layer to adjust the weights before the true output layer. Different from the multi-task learning criterion [24], which enables the "model" to process different tasks in the same time, the proposed MML tries to improve the performances of "outputs" to consider multiple metrics simultaneously.

**Table 2.** Performance comparisons of different models in terms of LSD, SSNR, STOI, and PESQ.

| SNR (dB) | DNN-baseline | | | | RI-DNN ($\alpha=1, \beta=0$) | | | | RI-CNN ($\alpha=1, \beta=0$) | | | |
|---|---|---|---|---|---|---|---|---|---|---|---|---|
| | LSD | SSNR | STOI | PESQ | LSD | SSNR | STOI | PESQ | LSD | SSNR | STOI | PESQ |
| 12 | **3.115** | -0.229 | 0.814 | 2.334 | 3.761 | 2.149 | 0.851 | 2.643 | 3.604 | **3.042** | **0.886** | **2.741** |
| 6 | **3.404** | -1.243 | 0.778 | 2.140 | 3.936 | 1.113 | 0.817 | 2.404 | 3.844 | **1.975** | **0.850** | **2.525** |
| 0 | **3.747** | -2.802 | 0.717 | 1.866 | 4.200 | -0.454 | 0.750 | 2.088 | 4.150 | **0.450** | **0.783** | **2.233** |
| -6 | **4.114** | -4.974 | 0.626 | 1.609 | 4.521 | -2.745 | 0.645 | 1.778 | 4.491 | **-1.911** | **0.675** | **1.908** |
| -12 | **4.426** | -7.070 | 0.521 | 1.447 | 4.838 | -5.604 | 0.512 | 1.539 | 4.829 | **-4.990** | **0.537** | **1.638** |
| Avg | **3.761** | -3.264 | 0.691 | 1.879 | 4.251 | -1.108 | 0.715 | 2.090 | 4.183 | **-0.286** | **0.746** | **2.209** |

## 5. EXPERIMENTS

### 5.1. Experimental setups

In our experiments, the TIMIT corpus [25] was used to prepare the training and test sets. 600 utterances were randomly selected and corrupted with five noise types (Babble, Car, Jackhammer, Pink, and Street), at six SNR levels (-15 dB, -10 dB, -5 dB, 0 dB, 5 dB, and 10 dB). Another 100 randomly selected utterances were mixed to form the test set. To make experimental conditions more realistic, both the noise types and SNR levels of the training and test sets were mismatched. Thus, we intentionally adopted three other noise signals: (White Gaussian noise, a stationary noise) and (Engine, Baby cry, non-stationary noises), with another five SNR levels: -12 dB, -6 dB, 0 dB, 6 dB, and 12 dB to form the test set. All the results reported in Section 5.2 are averaged across the three noise types.

In this work, 257 dimensional ($L$=257) LPS (for the baseline) and RI spectrograms (514 dimensions in total, 257 for each of R and I spectrograms) were extracted from the speech waveforms as acoustic features. Mean and variance normalization was applied to the input feature vectors to make the training process more stable. The DNNs in this experiment had six hidden layers (each with 1000 nodes) with parametric rectified linear units (PReLUs) [26] as activation functions. CNN had four convolutional layers with padding (each layer consisted of 50 filters each with a filter size of 25x1) and two fully connected layers (each with 512 nodes). Both models are trained using adam [27] with batch normalization [28].

**Table 1.** SSNR scores by combining clean magnitude spectrograms with noisy phase.

| Input SNR (dB) | SSNR (dB) |
|---|---|
| 12 | 13.43 |
| 6 | 9.931 |
| 0 | 6.847 |
| -6 | 4.248 |
| -12 | 2.149 |

To evaluate the performance of different models, SSNR and LSD were used for evaluating signal differences in the time domain and the frequency domain, respectively. In addition, the perceptual evaluation of speech quality (PESQ) [29] and the short-time objective intelligibility (STOI) scores [30] were employed to evaluate the speech quality and intelligibility, respectively. Although these two metrics are not included in the designed objective function of our MML, we also report the results for completeness.

### 5.2. Experimental results

#### 5.2.1. Effect of phase in different SNR conditions

In this section, we intend to investigate whether the explanation made in Section 2 is reasonable. We adopted the clean magnitude spectrograms with noisy phase from different SNRs (-12 dB to 12 dB) to synthesize waveforms. Table.1 shows the average SSNR scores of the synthesized waveforms and verifies that using noisy phase in low SNR conditions degrades the signal more severely.

#### 5.2.2. Comparison of different models

To separately investigate the effects of enhancing RI spectrograms and MML, the model with $\beta$=0 during training is denoted as RI-DNN or RI-CNN, and CNN with multi-metrics learning is denoted as MML-CNN. We first compare the proposed RI-CNN with RI-DNN and the DNN-baseline, which only enhances the magnitude spectrogram. Table 2 shows the quantitative results of the average LSD, SSNR, STOI, and PESQ scores on the test set, among the three models. As expected, the DNN-baseline model can reach the lowest LSD score since it enhances the LPS directly (not through the reconstruction from RI spectrograms). However, in terms of the other three metrics, RI-DNN shows noticeable improvements compared to the baseline. This suggests that enhancing the log-power-spectrogram alone may not yield satisfactory results on multiple metrics [13, 31]. In addition, please note that the huge improvement of SSNR in RI-DNN verifies the argument that optimizing RI spectrograms is related to maximizing SSNR. The results obtained by RI-CNN can further outperform RI-DNN, implying the superior feature extraction ability of the CNN model, as reported in [4].

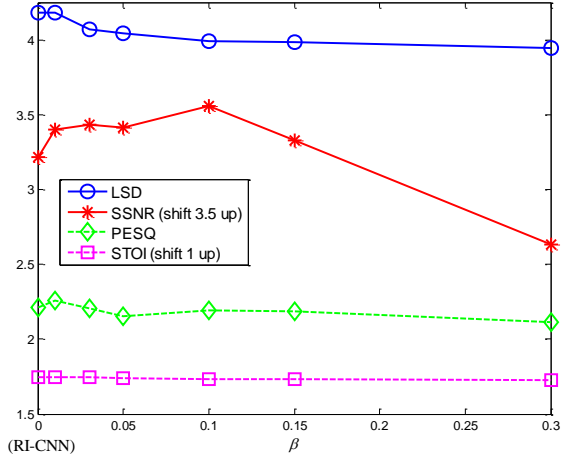

**Fig. 4.** Trends of the four metrics as function of $\beta$ with $\alpha = 1$ in MML-CNN. When $\beta = 0$, it degrades to RI-CNN.

We also try to only employee (5) or (7) (without (3)) as the objective function of DNN for waveform and LPS reconstruction, respectively (both results are not shown here due to the limited space). The enhanced results using (5) are similar to those of RI-DNN, because the only difference between (3) and (5) is just the linear transformation **F**. The results using (7) are similar to those of DNN-baseline, since they have the same objective function (even though (7) indirectly achieves this through the reconstruction from RI spectrograms).

*5.2.3. Results of MML*
To investigate the effects of MML, figure 4 shows the trends of the four metrics as function of $\beta$. Note that for clearly presenting all the trends in one figure, scores are linearly shifted to a similar range (SSNR is shifted up by 3.5, and STOI is shifted up by 1). The results show that increasing $\beta$ can effectively improve LSD as expected, while keeping STOI and PESQ roughly unchanged (we use dash line for the two metrics since they are not included in our objective function; the results are only for comparison purpose). Surprisingly, in the range of 0 to 0.1, increasing $\beta$ can also improve SSNR. This implies that, for small $\beta$, unlike the usual multi-objective optimization problem, the terms in (9) do not conflict with each other. Because the optimal solutions for all the terms in (9) are still the clean speech just represented in different forms. This SSNR improvement may be due to that RI-CNN estimated all the output nodes independently while MML made the estimated real and imaginary spectrograms influence each other as shown in Fig. 5. In other words, the RI-spectrograms have to cooperate with each other to produce a good estimation of LPS. This constraint may facilitate CNN better generalization and performance.

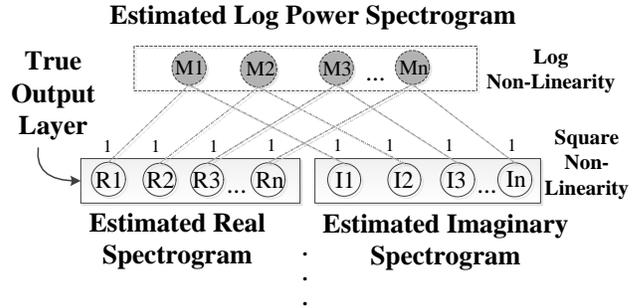

**Fig. 5.** Pseudo layer: applying the LPS reconstruction term in the objective function makes the estimated real and imaginary spectrograms influence each other.

## 6. CONCLUSIONS

The contribution of this paper is three-fold. First, we proposed a novel CNN-based speech enhancement model, which estimates clean RI spectrograms from noisy ones. The reconstructed RI spectrograms are then used to synthesize enhanced speech waveforms with more accurate phase information. Second, we derive an MML criterion that considered multiple metrics in the objective function. The main concept of MML is mainly based on other signal representation forms can be depicted as functions of RI spectrograms. Third, experimental results show that MML can simultaneously improve several objective metrics (LSD and SSNR) when $\beta$ is properly specified. The performance improvements can be explained by treating MML as adding constraints (pseudo layers) on the original objective function; this particular structure can enhance the generalization capability of the original model. In the future, we will investigate the integration of STOI and PESQ into the objective function to form a more complete MML. In addition, different forms of objective function (not simply weighted sum) used for MML will also be studied.